# Optimizing Structured Data Processing Through Robotic Process Automation

Vivek Bhardwaj[1], Ajit Noonia[1], Sandeep Chaurasia[1], Mukesh Kumar[2], Abdulnaser Rashid[3*], Mohamed Tahar Ben Othman[3]

[1] School of Computer Science and Engineering, Manipal University Jaipur, Jaipur 303007, India
[2] Department of Computer Applications, Chandigarh School of Business, Chandigarh Group of Colleges, Jhanjeri, Mohali 140307, India
[3] Department of Computer Science, College of Computer, Qassim University, Buraydah 51452, Saudi Arabia

Corresponding Author Email: arshied@qu.edu.sa

**ABSTRACT**

Robotic Process Automation (RPA) has emerged as a game-changing technology in data extraction, revolutionizing the way organizations process and analyze large volumes of documents such as invoices, purchase orders, and payment advices. This study investigates the use of RPA for structured data extraction and evaluates its advantages over manual processes. By comparing human-performed tasks with those executed by RPA software bots, we assess efficiency and accuracy in data extraction from invoices, focusing on the effectiveness of the RPA system. Through four distinct scenarios involving varying numbers of invoices, we measure efficiency in terms of time and effort required for task completion, as well as accuracy by comparing error rates between manual and RPA processes. Our findings highlight the significant efficiency gains achieved by RPA, with bots completing tasks in significantly less time compared to manual efforts across all cases. Moreover, the RPA system consistently achieves perfect accuracy, mitigating the risk of errors and enhancing process reliability. These results underscore the transformative potential of RPA in optimizing operational efficiency, reducing human labor costs, and improving overall business performance.

## 1. INTRODUCTION

By automating routine, rule-based processes that were previously completed by people, RPA is revolutionising the way businesses run. It entails the use of "bots," or software robots, to interface with various systems and applications to carry out tasks while imitating human behaviour [1, 2]. RPA makes use of machine learning (ML) and artificial intelligence (AI) to evaluate and comprehend the steps in a specific process. Once they are trained, the software robots can carry out those activities with greater accuracy, speed, and precision than people can, frequently. Finance, healthcare, manufacturing, customer service, and other fields and sectors can all benefit from the use of RPA. The idea behind RPA is to find repetitive, rule-based jobs that take a lot of time and are prone to mistakes when done manually. Data input, data validation, data extraction, report production, and other comparable procedures are examples of these duties (refer to Figure 1). Organizations can increase operational effectiveness, cut costs, improve accuracy, and speed up processing times by automating certain processes [3-5]. RPA can be applied in a variety of ways, from straightforward rule-based automation to more sophisticated intelligent automation. Automating predetermined tasks with simple RPA entails adhering to a set of rules or guidelines. Contrarily, intelligent automation combines cognitive tools like machine learning and natural language processing to provide bots the ability to manage unstructured data, make judgments, and adjust to changing circumstances. The non-invasive aspect of RPA, which can operate on top of current systems without requiring large changes to the underlying infrastructure, is one of its key advantages. It can be integrated with a wide range of software programmes, legacy systems, databases, websites, and more, enabling businesses to make the most of their current IT expenditures. Due to its ability to boost productivity, streamline processes, and free up employees to work on higher-value tasks, RPA is quickly gaining appeal. As RPA continues to influence the future of labour and corporate processes, academics and educators must keep up with its developments and ramifications.

RPA uses software robots to simulate human behaviour and communicate with a variety of software programmes and systems [5]. When it comes to data extraction, RPA bots may be taught to efficiently and accurately extract data from a variety of sources, including databases, websites, emails, and documents. RPA dramatically lowers the chance of errors, improves data quality, and boosts productivity by removing manual data entry and extraction operations. RPA also gives businesses the ability to manage difficult data extraction scenarios involving structured, semi-structured, and unstructured data. RPA bots can intelligently extract pertinent information and transform it into organised formats for additional analysis since they are able to interpret and comprehend various data formats. With the help of this capacity, firms may have access to insightful data and take faster, more precise data-driven decisions. The advantages of RPA in data extraction go beyond boosted productivity and enhanced data quality. This lowers total operational expenses by allowing firms to reallocate human resources to more strategic and value-added duties. Continuous data extraction and processing is possible with RPA bots, which improve operational agility and responsiveness. RPA implementation for data extraction, however, necessitates thoughtful planning and study. Companies need to determine appropriate use cases, create solid automated procedures, and guarantee data security and compliance. To resolve any possible problems and improve performance, RPA processes must also be continuously monitored and maintained.

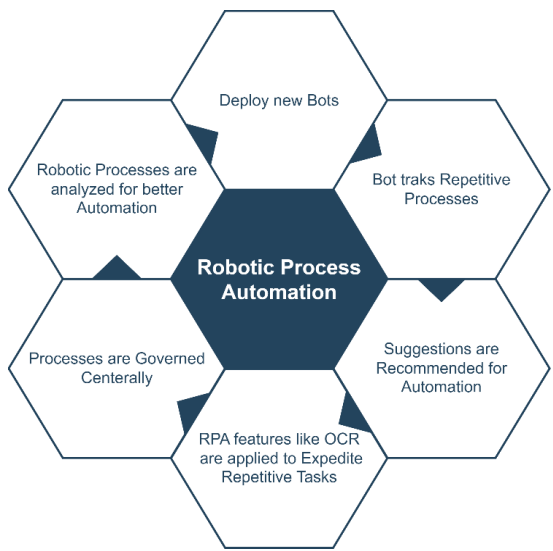

**Figure 1.** RPA for repetitive processes

RPA is enhancing efficiency and reducing costs across various industries. In financial services, RPA automates data entry, transaction processing, compliance checks, and reporting, significantly reducing processing times and errors. In healthcare, it manages patient records, schedules appointments, and processes billing and claims, allowing healthcare professionals to focus on patient care. In manufacturing, RPA is used for inventory management, supply chain operations, and quality control, helping to reduce production downtime and optimize logistics. In retail, it automates order processing, customer service, and inventory management, leading to faster order fulfillment and improved customer satisfaction. In telecommunications, RPA automates customer service, network management, and billing, efficiently handling large volumes of inquiries and improving service.

This study investigates the effectiveness of RPA in extracting data from invoices by comparing its performance to manual extraction methods. The research aims to provide insights into the specific benefits of RPA, including its efficiency in processing time, accuracy in data extraction, and overall impact on operational practices. To address these objectives, the following research questions guide this study:

**Efficiency in data extraction:** How does the time required for data extraction from invoices using RPA compared to the time required for manual extraction across varying numbers of invoices?

**Accuracy in data extraction:** How does the accuracy of data extraction from invoices using RPA compared to the accuracy of manual extraction?

## 2. LITERATURE REVIEW

RPA has indeed ignited a significant volume of literature spanning diverse domains, providing valuable insights into its applications, impacts, and future potential. Here's an overview of the literature landscape:

One can find papers on the potential for using IT solutions for long-term business continuity management in the literature [3, 6]. The present pandemic and lockdown time are also considered by several authors, who offer reflection on how the pandemic exposed the fragility of digitally immature organizations, education, employment, and life. In various fields, including information technology, human resources, telecommunication, finance, real estate management, insurance, education, legal services, banking, and logistics, automated software robots are becoming more and more common [7, 8].

To analyse the use of RPA within a BPO unit, Aguirre and Rodriguez conducted a case study [9]. In their investigation, they discovered that productivity increased, and processing times dropped when RPA was utilised for a few of the organization's back-office tasks. Perez-Arriaga et al. [10] give Table Organization (TAO). It is a method for finding, extracting, and organising data from tables in PDF documents automatically. TAO recognises and extracts table information from documents using layout heuristics and k-nearest neighbour processing. This technique improves the representation of the data extracted from PDF tables. TAO's performance is comparable to that of earlier table extraction approaches, but it has more flexibility and more robustness when tested with different document layouts.

Day-Yang, Shou-Wei, and Tzu-Chuan stressed that the use of such technologies is causing a change in how businesses work, and that this change has as its result the close fusion of digital technology and business processes. As a result of this integration, new models for how these institutions operate are created, with digital technology at their centre [11]. Define the software robot concept as a starting point for an investigation of the use of RPA tools as one of the strategies for executing digital transformation. Willcocks and Lacity [12] claim that it is software that automates specific human tasks that are performed as part of a specified business process, most frequently by accurately replicating (mimicking) them. According to this method, a robot is not recognised by a technological gadget and does not have the ability to move like a human.

Aldoseri et al. [13] discussed the integration of automation and AI-driven processes represents a transformative shift in how businesses operate, offering substantial benefits in efficiency, decision-making, and innovation. However, realizing these benefits requires addressing challenges related to technology implementation, change management, and cultural adaptation. Future research should focus on industry-specific applications and detailed case studies to provide tailored insights and strategies, ensuring that organizations can successfully navigate the complexities of AI-powered digital transformation.

Huang et al. [14] demonstrates the successful application of Lean Six Sigma DMAIC and RPA in streamlining the medical expense claims process under Taiwan's NHI system, significantly reducing process time and enhancing efficiency. By reallocating human resources to more valuable tasks, healthcare institutions can offer more comprehensive and patient-centric services, showcasing a real-world example of Lean digital transformation in the healthcare sector. Future research should explore the broader implementation of these approaches in various healthcare settings to further validate their benefits.

Bhadra et al. [15] underscores the transformative potential of integrating CIoT and RPA to drive innovations in industrial automation, enhancing situational awareness and autonomous operations in the industry 4.0 era. By presenting unique architectural semantics and compelling use cases, it highlights the design rationale for next-generation cognitive enterprise systems, paving the way for future research on autonomous systems enabled by this convergence.

## 3. RPA VS BPM

Business process management and robotic process automation are two different but complimentary methods for automating processes in enterprises. Despite their commonalities, they have diverse focuses and are used for various things.

**Scope:** RPA often automates repetitive and rule-based tasks by focusing on certain tasks or activities within a process. Without significantly altering the general process flow, it concentrates on automating particular steps inside already-existing processes. BPM, on the other hand, is concerned with the overall management and optimization of business processes from beginning to end. Whole processes are analysed, modelled, designed, and monitored as a part of this process, which frequently aims to increase effectiveness, efficiency, and agility [16].

**Automation level:** RPA is best used to automate repetitive, manual processes that entail interacting with software programmes. RPA robots mimic human behaviour and carry out activities using a user interface (UI). For jobs involving organised and semi-structured data, it is effective. Contrarily, BPM focuses on managing and orchestrating processes. Automation may be a part of it, but it also takes into account other things like process design, workflow management, teamwork, and integration with other systems and stakeholders [17].

**Process complexity:** RPA works well for automating repeated operations that are generally easy to perform. It is frequently used for data entry, data verification, and report production duties [18]. Complex processes with several steps, decision points, and interactions between numerous systems and stakeholders are better managed with BPM. It permits modelling and process optimization for whole workflows, including both automated and human tasks.

RPA implementations are often quicker and more agile than BPM implementations in terms of implementation strategy. Without making significant changes to the current IT infrastructure, individual processes can be swiftly deployed and automated using RPA bots. On the other hand, BPM implementations take a more strategic, all-encompassing approach. They necessitate process modification and analysis, and frequently involve integrating different systems and stakeholders. BPM implementations could take longer, but they might also affect the organisation more broadly. RPA and BPM can be combined in the real world to maximise automation and process improvement. RPA can be used to automate repetitive operations and lessen manual labour within steps of a BPM-managed process. RPA gives the tactical automation tools to speed up certain process tasks, whereas BPM provides the overall framework for process management and optimization.

## 4. BENEFITS AND COMPARATIVE ANALYSIS OF RPA TOOLS

RPA is revolutionizing how businesses operate by automating repetitive tasks previously performed by humans. Here's a detailed breakdown of the benefits it offers to organizations:

RPA streamlines operations by automating mundane, repetitive tasks, freeing up employees to focus on higher-value activities. This efficiency boost increases productivity and throughput without the need to expand the workforce. Robots execute tasks with precision, significantly reducing errors and enhancing data quality. This is particularly advantageous for tasks requiring meticulous attention to detail, such as data entry and processing, ensuring reliable and error-free outcomes.

RPA leads to substantial cost reductions by automating tasks that were previously labor-intensive. Organizations can save on labor costs and operational expenses while achieving higher output levels, contributing to improved financial performance and competitiveness. RPA solutions are highly scalable and adaptable to varying workloads and business demands. Whether experiencing surges in activity or scaling down operations, organizations can easily adjust their robotic workforce without significant additional investment, ensuring optimal resource allocation.

Robots execute tasks according to predefined rules and regulations, ensuring strict adherence to company policies and industry standards. This minimizes the risk of compliance breaches and associated penalties, bolstering organizational integrity and reputation. By automating repetitive tasks, such as order processing and customer inquiries, organizations can streamline operations, reduce response times, and enhance overall customer satisfaction. Improved efficiency and accuracy lead to smoother transactions and interactions, fostering loyalty and positive brand perception.

RPA generates valuable data insights during task execution, providing organizations with actionable analytics for informed decision-making and process optimization. This data-driven approach empowers organizations to identify trends, opportunities, and areas for improvement, driving strategic business growth. RPA implementations typically yield a rapid return on investment (ROI) due to their quick deployment and tangible benefits. Organizations can realize cost savings, productivity gains, and efficiency improvements in a relatively short period, accelerating the realization of ROI and maximizing the value of their investment.

In summary, RPA empowers organizations to operate more efficiently, reduce costs, ensure compliance, enhance customer experience, and leverage data-driven insights, ultimately driving sustainable growth and competitiveness in today's dynamic business landscape.

### 4.1 RPA tools

Tools for automating repetitive operations and streamlining corporate processes are known as robotic process automation (RPA) tools. Comparison of UiPath, Automation Anywhere, Blue Prism, Work Fusion, Kofax, and Pega are shown in the Table 1. UiPath is appropriate for sophisticated automation scenarios because it has a visual workflow builder and supports many applications. A user-friendly interface is offered by Automation Anywhere, which enables desktop and web automation. Secure credential management and a centralised control room are two features that Blue Prism, which specialises in enterprise-grade automation, provides. By automating manual operations and allocating human resources to more worthwhile duties, these RPA systems help firms boost efficiency, accuracy, and production [19, 20].

**Table 1.** Overview of leading RPA tools

| RPA Tool | Key Features | Supported Platforms | Ease of Use | Integration Capabilities | AI/ML Capabilities | Community Support |
|---|---|---|---|---|---|---|
| UiPath | Drag-and-drop workflow designer | Windows, Web, Citrix | User-friendly | Extensive integrations | Built-in AI | Active community |
| Automation Anywhere | Process recording, pre-built components | Windows, Web, Citrix | Intuitive | Wide range of integrations | Advanced AI | Robust community |
| Blue Prism | Visual process designer, centralized management | Windows, Web, Citrix | Moderate | Strong integration capabilities | Limited AI | Support available |
| Work Fusion | AI-powered automation, data extraction | Windows, Web, Citrix | Moderate | Advanced integration capabilities | Advanced AI | Support available |
| Kofax | Data capture, high-volume processing | Windows, Web, Citrix | Moderate | Extensive integrations | Limited AI | Support available |
| Pega | BPM integration, case management | Windows, Web, Citrix | Moderate | Robust integration capabilities | Advanced AI | Active community |

## 5. UIPATH

The client-server architecture used by UiPath makes it possible to create, build, and run automation processes [21]. The UiPath Studio, UiPath Robot, and UiPath Orchestrator are the three core parts of the architecture which is shown in Figure 2.

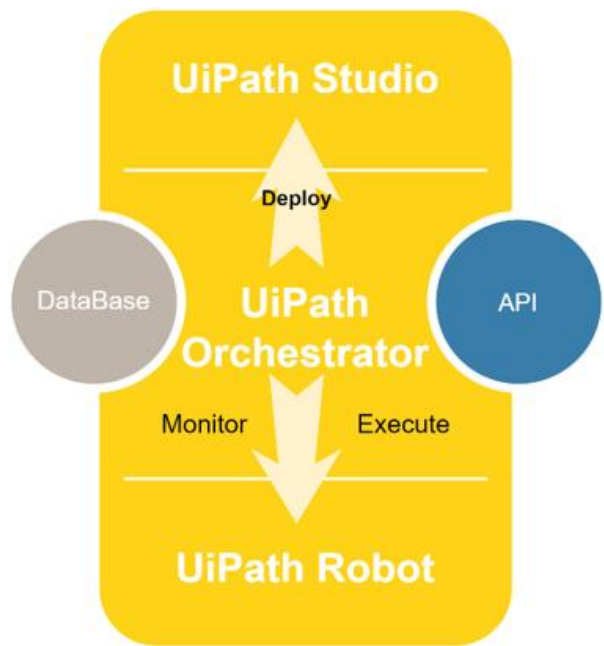

**Figure 2.** UiPath architecture

UiPath Studio: The platform for developing automation workflows is provided by this component. To develop and edit workflows, it offers a visual interface with drag-and-drop capabilities. Users of UiPath Studio can specify the order of activities, work with data, and add different activities from the activity library. It also offers logging features, debugging tools, and the capacity to communicate with programmes and systems using various automation strategies [19].

UiPath Robot: The automation workflows made in UiPath Studio are carried out by the UiPath Robot component. It is in charge of interfacing with programmes, systems, and UI elements and can be installed on single workstations or virtual environments. The Robot can handle both attended and unattended automation settings and executes workflows by adhering to its defined instructions. To accept instructions, report progress, and fetch new automation tasks, it interfaces with UiPath Orchestrator.

UiPath Orchestrator: A centralised web-based platform called UiPath Orchestrator serves as the command centre for coordinating and keeping track of automation activities. It offers a centralised interface for managing automation workflows across various devices and environments. Asset management, scheduling, queue management, user and role administration, and reporting are functions that Orchestrator provides. Version control, deployment, and auditing of automated processes are also made easier. Through orchestrator, automation may be coordinated and scaled across a company, resulting in effective administration and oversight of robots.

UiPath Architecture additionally includes other elements including UiPath Assistant, which offers a user-friendly interface for engaging with attended robots, and UiPath AI Center, which makes it easier to integrate and deploy AI and machine learning models within automation processes. A reliable foundation for creating, deploying, and managing automation at scale is offered by UiPath's architecture. It allows for seamless developer collaboration, robot execution of automation tasks, and Orchestrator-based centralised control and monitoring.

These elements work together in the following ways: Automation workflows are created in UiPath Studio and then published to Orchestrator. UiPath Robot implements the published workflows after receiving instructions and automation tasks from Orchestrator. Workflow deployment to robots is managed by Orchestrator, which also schedules execution and offers centralised monitoring and reporting.

Users can initiate and keep an eye on automation operations using the UiPath Assistant user interface, which enables human engagement with attended robots. The integration of AI and machine learning models into automation workflows is made possible by UiPath AI Center, boosting the capabilities of the automation processes.

## 6. METHODLOGY AND RESULTS

The ability to accurately extract data from documents like invoices, purchase orders, payment advices, and other bespoke documents is one of the most crucial aspects of automating business activities.

The information included in these documents must be manually read and comprehended in a laborious manner in a typical workflow, which adds time and expense for any organisation. Figure 3 shows the steps followed for extracting the structured data from invoices using RPA. Organizations can now unwind and complete the time-consuming and stressful jobs with a single click thanks to the integration of Intelligent RPA and Document Information Extraction service, as demonstrated in the image below.

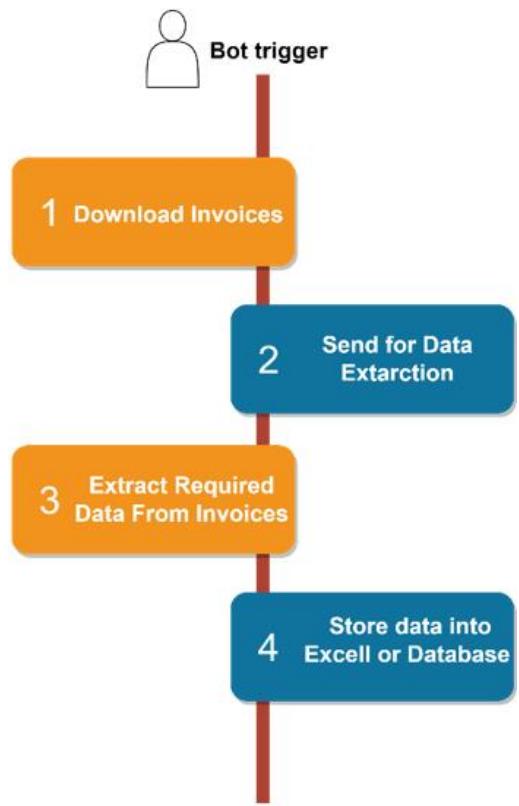

**Figure 3.** Methodology for extracting structured data

In the analysis and evaluation of the manual and RPA processes, we compared human-performed tasks with those executed by RPA software bots. The focus was on measuring efficiency and accuracy in extracting data from processed invoices, aiming to assess the effectiveness of the RPA system.

UiPath Studio was selected for this study due to its advanced capabilities in developing and deploying RPA workflows tailored for invoice processing tasks. This version was chosen to leverage the latest features and improvements in the software, ensuring optimal performance and functionality for the study.

Visual Workflow Designer: UiPath Studio includes an intuitive visual workflow designer that allows users to create and manage automation processes through a user-friendly, drag-and-drop interface. This feature is essential for designing complex workflows without requiring extensive coding expertise.

UiPath.PDF Activities Dependency: The UiPath.PDF Activities package was integrated into UiPath Studio to enhance the extraction and processing of data from PDF invoices. This package provides specialized activities for reading and extracting data from PDF documents, crucial for handling the varied formats and layouts of invoice data.

UiPath.Database.Activities: In our study, we utilized this package for interacting with databases. We employed activities such as Execute Query, Execute Non-Query, and Insert, Update, and Delete to store and manage data efficiently within the database.

The manual data extraction process involved five operators: two data entry personnel who transcribed invoice details into the system, one verification specialist who ensured data accuracy, one data reviewer who checked for completeness, and one quality assurance analyst who audited the entire process for errors. Adherence to detailed Standard Operating Procedures (SOPs) and accuracy checks were critical in maintaining data integrity and consistency. This manual process serves as a baseline for evaluating the efficiency and accuracy of RPA.

To conduct a comprehensive evaluation, we designed five distinct scenarios, each involving a different number of invoices:

Case 1 (10 invoices): In this scenario, we processed a set of 10 invoices using both manual and RPA methods to extract relevant data.

Case 2 (15 invoices): Similarly, we expanded the scope to 15 invoices to observe how the performance of both manual and RPA processes scaled with increased workload.

Case 3 (20 invoices): This scenario involved processing 20 invoices, providing further insights into the scalability and efficiency of the RPA system across a larger dataset.

Case 4 (25 invoices): We examined the performance of both methods in handling 25 invoices, aiming to assess the upper limits of scalability and the potential impact on efficiency and accuracy.

In Case 5, the evaluation extends to processing a set of 30 invoices using both manual and RPA methods. This expanded scenario allows for a comprehensive assessment of the scalability, efficiency, and accuracy of both approaches when handling a larger workload.

By increasing the number of invoices to 30, we aim to gain deeper insights into how both manual and RPA processes perform as the workload grows. This includes observing any potential bottlenecks or limitations in scalability, as well as assessing the impact on efficiency and accuracy.

For each case, we meticulously measured and compared several key metrics:

**Efficiency:** This metric quantifies the time and effort required to complete the task. We compared the time taken for humans to manually extract data from invoices against the time taken by RPA bots to perform the same task.

**Accuracy:** Accuracy measures the precision and reliability of data extraction. We compared the error rates between human-performed tasks and those executed by RPA bots, focusing on discrepancies or mistakes in the extracted data.

By analysing the data collected from these five scenarios,

we gained valuable insights into the performance of the RPA system compared to manual processes. These insights allow us to make informed decisions regarding the implementation and optimization of RPA solutions within the organization, highlighting areas of strength and opportunities for improvement.

Figure 4 illustrates the stark contrast in processing speed between Manual and RPA Process.

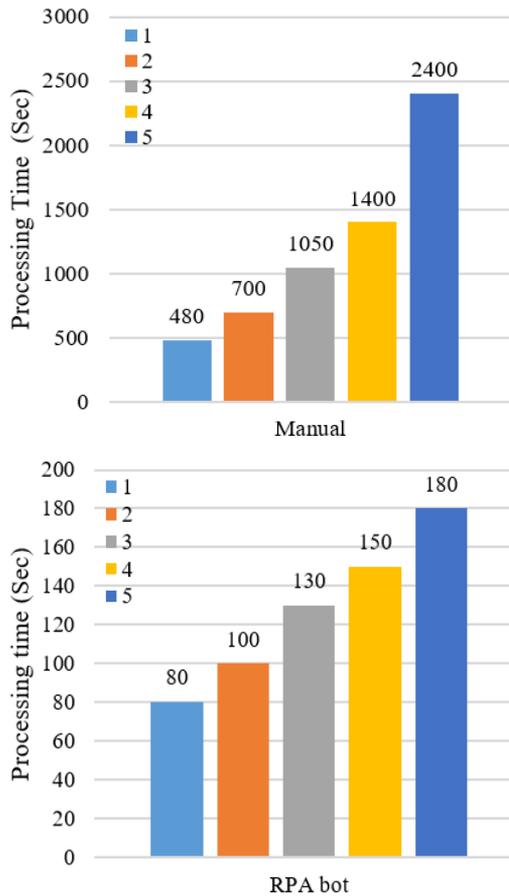

**Figure 4.** Processing time of Manual v/s RPA process

The results from the comparison between manual task completion and (RPA performance demonstrate a clear advantage for RPA across all cases. In each instance, the RPA bot completed the task in significantly less time compared to manual efforts.

Case 1 showed an exceptional efficiency gain, with the RPA bot completing the task 6 times faster than a human. This highlights the suitability of RPA for highly repetitive tasks, where its speed and accuracy can yield substantial time savings and productivity gains.

Similarly, in Case 2, the RPA bot outperformed manual completion by completing the task 7 times faster. This consistent pattern of efficiency gains is reiterated in Case 3, where the RPA bot completed the task 8 times faster than the manual process.

Even in more time-consuming tasks, such as Case 4 and Case 5, the RPA bot demonstrated its efficiency by completing the tasks in a fraction of the time required by a human. Despite the task complexities and longer durations, the RPA bot maintained its speed advantage, completing tasks in 9.3 and 13.3 times less time than manual efforts, respectively.

Overall, these results underscore the transformative potential of RPA in optimizing operational efficiency and reducing human labor costs. By automating repetitive and rule-based tasks, RPA enables organizations to streamline processes, improve productivity, and allocate human resources to more strategic and value-added activities. This highlights the strategic significance of RPA adoption in enhancing organizational agility and competitiveness in the digital era.

Figure 5 illustrates the contrast in processing accuracy between Manual and RPA Process. From the results we found that the RPA works with higher accuracy than the manual process. The comparison of accuracy between manual task completion and Robotic Process Automation (RPA) performance across the five cases illustrates a clear advantage for the RPA bot, which consistently achieves perfect accuracy (100%). Meanwhile, the manual process exhibits slight variations in accuracy ranging from 80% to 98%.

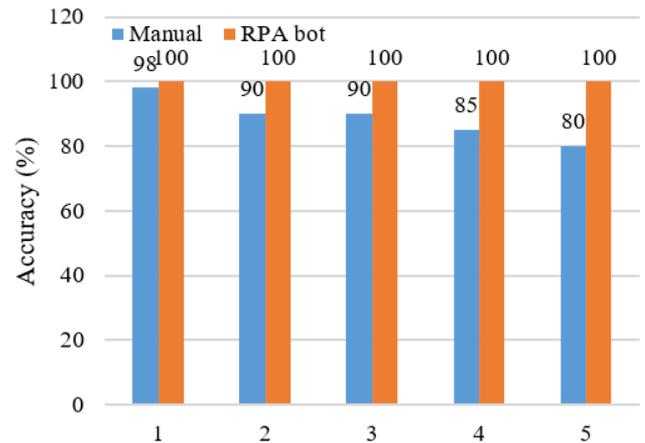

**Figure 5.** Processing accuracy of Manual v/s RPA process

In Case 1, despite the manual process achieving a commendable accuracy rate of 98%, the RPA bot stands out by delivering flawless performance, indicating its capability to execute tasks without errors.

Similarly, in Case 2, where the manual accuracy drops slightly to 90%, the RPA bot maintains its perfect accuracy, showcasing its reliability in ensuring error-free execution.

Continuing through the cases, the manual process maintains an accuracy of 90% in Case 3, 85% in Case 4, and 80% in Case 5. In contrast, the RPA bot consistently achieves perfect accuracy across all scenarios, regardless of the fluctuations in manual performance. This consistency highlights the reliability and precision of RPA in executing tasks without errors, ensuring consistent and high-quality outcomes. Moreover, the perfect accuracy achieved by the RPA bot suggests a reduced risk of errors, rework, or compliance issues, further emphasizing the value of automation in enhancing process reliability and quality assurance.

Overall, the results underscore the transformative potential of RPA in optimizing operational efficiency and minimizing risks associated with human error, making it a valuable asset for organizations aiming to streamline processes and improve overall performance.

The scalability of RPA systems significantly enhances their ability to manage large volumes of invoices efficiently. RPA's design allows it to handle increasing workloads without compromising performance. As demonstrated in various cases, RPA systems can process invoices substantially faster than manual methods, with efficiency improvements ranging from 6 to 13 times quicker. This scalability ensures that businesses can accommodate growing transaction volumes seamlessly, maintaining high-speed processing and accuracy as invoice numbers rise.

Moreover, the perfect accuracy achieved by RPA systems has profound implications for businesses. By eliminating errors, RPA enhances the reliability of data used for decision-making, leading to more informed and strategic choices. This accuracy supports stringent compliance with regulatory standards, as the risk of human error and associated penalties is minimized. Additionally, RPA's flawless performance streamlines operations, reducing the need for rework and improving overall efficiency. Integrating intelligent automation further enhances decision-making by leveraging data-driven insights and predictive analytics, enabling businesses to anticipate trends, optimize processes, and drive continuous improvement. Thus, RPA not only scales effectively with growing workloads but also significantly enhances operational precision and strategic decision-making.

## 7. CONCLUSIONS

In conclusion, our study demonstrates the remarkable benefits of integrating Robotic Process Automation (RPA) into data extraction processes. By automating repetitive and rule-based tasks traditionally performed by humans, RPA significantly enhances efficiency and accuracy, leading to substantial time savings and productivity gains for organizations. The consistent performance of RPA in achieving perfect accuracy underscores its reliability and precision, mitigating the risk of errors and ensuring high-quality outcomes. These findings emphasize the strategic significance of RPA adoption in enhancing organizational agility, competitiveness, and overall operational excellence in the digital era. Moving forward, organizations should capitalize on the transformative potential of RPA to streamline processes, optimize resource allocation, and drive sustainable growth in an increasingly data-driven business landscape.

## ACKNOWLEDGMENT

The Researchers would like to thank the Deanship of Graduate Studies and Scientific Research at Qassim University for financial support (QU-APC-2024-9/1).